\documentclass{article}
\usepackage[a4paper, margin=1in]{geometry} 
\usepackage{times}
\usepackage{graphicx}
\usepackage{booktabs}
\usepackage{xcolor}
\usepackage{xspace}    
\usepackage{textcomp}  
\usepackage{amsmath}   

\newcommand{\sysname}{\mbox{\texttt{FASTGEN}}\xspace}
\newcommand{\gt}{\mbox{\texttt{gt}}\xspace}
\newcommand{\rbr}{\mbox{\texttt{direct-gen}}\xspace}

\usepackage{amsmath,amsfonts,bm}

\def\eqref#1{equation~\ref{#1}}

\def\1{\bm{1}}

\DeclareMathAlphabet{\mathsfit}{\encodingdefault}{\sfdefault}{m}{sl}
\SetMathAlphabet{\mathsfit}{bold}{\encodingdefault}{\sfdefault}{bx}{n}

\usepackage{hyperref}
\usepackage{url}

\title{FASTGEN: Fast and Cost-Effective Synthetic Tabular Data Generation with LLMs}

\author{
Anh Nguyen, Sam Schafft, Nicholas Hale \& John Alfaro \\
Trillion Technology Solutions, Inc. \\
1950 Roland Clarke Pl \#410, Reston, VA 20191, USA \\
\texttt{\{anguyen, sschafft, nhale, jalfaro\}@ttsiglobal.com}
}

\begin{document}


\maketitle

\begin{abstract}
Synthetic data generation has emerged as an invaluable solution in scenarios where real-world data collection and usage are limited by cost and scarcity. Large language models (LLMs) have demonstrated remarkable capabilities in producing high-fidelity, domain-relevant samples across various fields. However, existing approaches that directly use LLMs to generate each record individually impose prohibitive time and cost burdens, particularly when large volumes of synthetic data are required. In this work, we propose a fast, cost-effective method for realistic tabular data synthesis that leverages LLMs to infer and encode each field’s distribution into a reusable sampling script. By automatically classifying fields into numerical, categorical, or free-text types, the LLM generates distribution-based scripts that can efficiently produce diverse, realistic datasets at scale without continuous model inference. Experimental results show that our approach outperforms traditional direct methods in both diversity and data realism, substantially reducing the burden of high volume synthetic data generation. We plan to apply this methodology to accelerate testing in production pipelines, thereby shortening development cycles and improving overall system efficiency. We believe our insights and lessons learned will aid researchers and practitioners seeking scalable, cost effective solutions for synthetic data generation.
\end{abstract}

\section{Introduction}

Synthetic data has been widely used in various applications where real data is scarce and expensive. Both research and industry have extensively focused on different types of synthetic data, such as images, articles \cite{brown2020language} and source code \cite{chen2021evaluating}. Among them, tabular data is one of the most commonly used formats in practical domains such as healthcare  \cite{choi2017generating, jung2024enhancing}, due to its structured format that enables automatic processing while remaining human readable.

The development of LLMs has introduced new opportunities for synthetic data generation. The ability of LLMs to memorize and generalize patterns from vast amounts of data enables the generation of realistic synthetic data across a wide range of fields \cite{bommasani2021opportunities}. To further improve output quality, recent work has explored fine-tuning LLMs to generate high quality, domain relevant data \cite{yu2023large}. In the majority of these works, LLMs have been used to directly generate individual records \cite{yu2023large, borisov2022language, ugare2024itergen, jung2024enhancing, guo2024generative}. However, direct generation is difficult to scale due to the time and cost of hosting and running LLMs. It becomes infeasible when large volumes of synthetic data are needed within tight resource constraints.

In this report, we propose Fast Automated Synthetic Tabular GENeration (\sysname), a fast and cost-effective approach for realistic synthetic data generation. Rather than asking an LLM to emit every record, we invoke the LLM to infer the underlying distribution using metadata descriptions provided by users. The LLM then returns sampling scripts (e.g., in Python) that can produce datasets of arbitrary size. This approach is fast and inexpensive, enabling the efficient production of large-scale synthetic datasets without continuous LLM inference, while still providing acceptable realism.

Achieving this goal is non-trivial, as the diverse nature of data domains requires the generation script to accommodate a wide range of data types. To handle this diversity, we categorize fields into three common types, numerical, categorical, and free-text. We then instruct the LLM to infer the distribution of each field accordingly. Our evaluation demonstrates that this distribution-based strategy produces more diverse and realistic field-wise data than traditional direct generation methods, while offering substantial improvements in efficiency.

At Trillion, we plan to apply our generation methodology to synthesize test datasets for production systems. Users provide metadata descriptions of the desired data, and the system automatically synthesizes inputs for downstream applications. By reducing data generation bottlenecks during testing, \sysname accelerates development cycles and improves overall efficiency. We also share insights and lessons learned from implementing our generation method, which we believe will benefit others looking to leverage LLMs for scalable data generation.

\section{Related Works}
Early work on data generation relied on expert domain knowledge \cite{faker, namegenerator} to craft generation rules or used statistical tools to captured distribution of numerical fields \cite{patki2016synthetic}. This approach works well for certain attributes such as name, location, and address, yet requires significant manual effort to keep up with the complexity and diversity of real-world data. Another line of research leverages deep networks, such as Generative Adversarial Networks \cite{goodfellow2020generative} and Variational Autoencoders \cite{kingma2013auto}, or small sized transformer architectures, e.g., GPT-2 or T5 \cite{radford2019language, raffel2020exploring}, to produce synthetic data. Despite these efforts, text output from earlier works lacked variation and was less coherent \cite{semeniuta2017hybrid} compared to more recent models.

Large language models are a boon for data generation. The expressive capability of LLMs has been noted and applied to fields where data collection for long-tail distributions is difficult \cite{kumichev2024medsyn, nia2025transforming}. In the majority of previous works, an LLM-assisted approach has been used to generate data record by record \cite{yu2023large, borisov2022language, ugare2024itergen, jung2024enhancing, guo2024generative}. This process is computationally expensive and time-prohibitive, making it impractical for applications that require generating a large number of records in a short period.

A common strategy to mitigate the computational overhead of LLM-based data generation is knowledge distillation, where a smaller model is trained to approximate the behavior of a larger model \cite{sanh2019distilbert}. However, this approach lacks flexibility, as the distillation process must be repeated whenever adaptation to a new domain is necessary.

In contrast to prior works, \sysname directly leverages the expressive power of large LLMs to generate diverse and high fidelity synthetic data while satisfying stringent latency constraints.

\section{Sampling Script Generation}
\begin{figure}[h!]
    \centering
    \includegraphics[width=.45\textwidth]{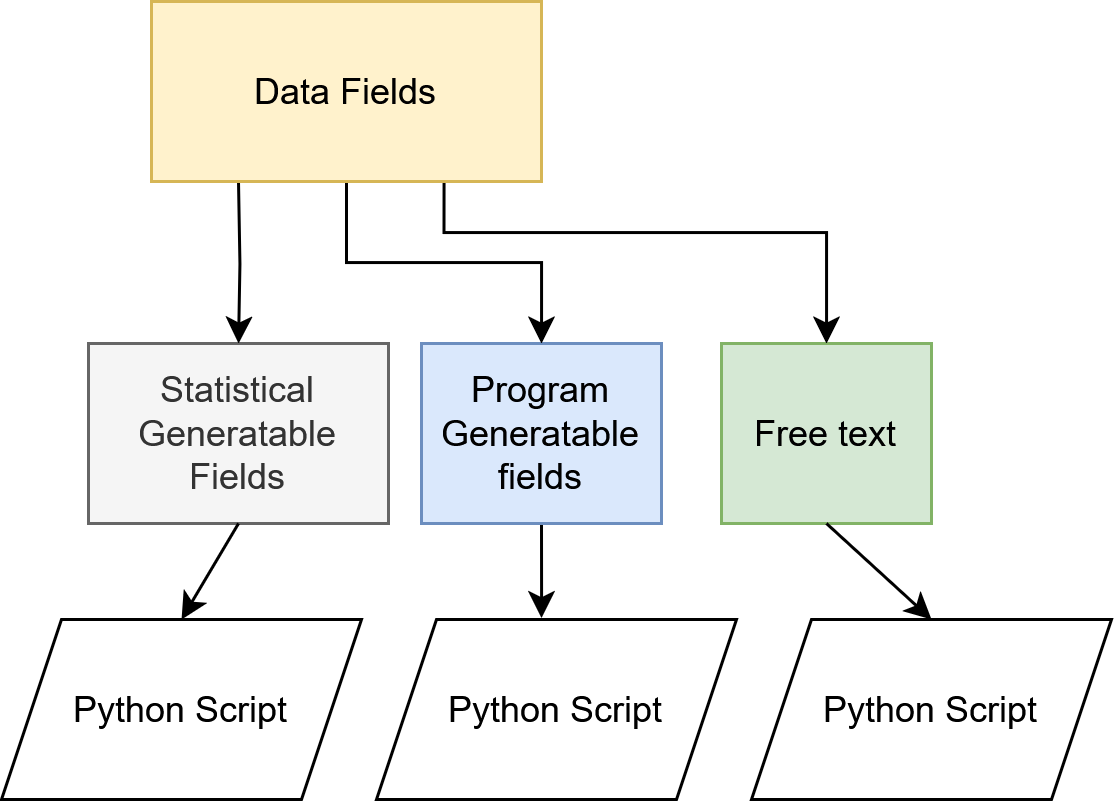}
    \caption{Overview of \sysname's approach for distribution estimation and script generation}
    \label{fig:method_exec_gen}
\end{figure}

Our method first analyzes the underlying distribution of each data field before attempting generation. As illustrated in \autoref{fig:method_exec_gen}, this process consists of three main steps. First, we instruct the LLM to examine the meta descriptions of each tabular field, thereby capturing the statistical properties of the data. Second, the LLM uses the distributional insights to produce scripts for data generation. Third, we run the scripts to obtain the actual synthetic data.

\subsection{Preprocessing}
Metadata descriptions provided by the original authors often omit essential information about the fields. These descriptions are typically intended to be brief summaries and were not designed with data generation tasks in mind. The following is an example of a field description for the \texttt{age} field from the Sick dataset:

\begin{quote}
\texttt{"age", "continuous", "Age of the individual"}
\end{quote}

This description is overly simplistic, which can lead to ambiguity and cause the generated data to fall short of users' expectations. To provide more suitable metadata descriptions, we prompt a large language model (LLM) to take on the role of a dataset curator and generate enriched metadata. The LLM is instructed to analyze $s$ randomly sampled ground truth values from each field and produce a concise description tailored for data generation tasks. This step is optional but recommended for users who want the generated data to meet specific requirements.

\subsection{Identify Distribution}

Determining underlying distributions is a challenging task, as fields may contain various data types. Our systematic approach is to instruct the LLM to categorize fields into three groups, numerical, categorical, and free-text, based on the metadata description of fields. An LLM may gain different insights depending on the group a field is assigned to.

\begin{itemize}
    \item \textbf{Numerical fields} are often modeled by well-defined distributions, e.g., uniform, normal, Poisson. These fields may include integers, floating-point numbers, or derived metrics such as ages, amounts, counts, currency.

    \item \textbf{Categorical fields} follow discrete distributions, where each category is associated with a probability. Enumerating all possible categories may be impractical due to token limitations or the sheer number of unique values. In some cases, category inference is challenging, as it relies heavily on field metadata descriptions. For instance, values like project codes or organization codes depend on domain specific business intelligence and cannot be inferred solely from the knowledge embedded in an LLM’s parameters.

    \item \textbf{Free text fields} contain unstructured data with no well-defined underlying distribution. In most cases, an LLM can only infer a limited set of frequent values. Due to this limitation, free-text fields present the greatest challenge.
    
\end{itemize}

\subsection{Script Generation}
We instruct the LLM to adopt different generation strategies tailored to the distribution characteristics.

\begin{itemize}
    \item \textbf{Numerical fields.} The LLM is instructed to estimate the distribution type and parameters (e.g., normal with mean and variance, uniform with min and max) and generate a script that samples data accordingly.
    
    \item \textbf{Categorical fields.}
    We instruct the LLM to identify the most frequent $k$ categories, assigning probabilities where possible. If no probabilities are provided, we default to a uniform distribution. A key limitation is that any rare yet semantically important categories may be dropped in the process, reducing the fidelity of the generated data.

    \item \textbf{Free text fields.}
    Rather than imposing a strict probability model, we instruct the LLM to generate random text that aligns with field-specific metadata. This strategy has two key benefits, it leverages the LLM’s creativity to ensure sufficient flexibility in accommodating diverse domain requirements, and it relies on the LLM's generative capabilities to produce realistic data.

Following these guidelines, the LLM produces a field-specific script for each field.
    
\end{itemize}



\subsection{Data Generation}
Once all field-specific scripts have been generated, we validate and execute field-specific scripts to verify potential errors. We perform an automated retry that prompts the LLM for corrections up to $n$ times. If the code remains non-executable after $n$ attempts, we return an empty or placeholder result for that problematic field. Field-specific scripts are then combined into a unified data generation script. The unified script is executed to create the final synthetic dataset

\section{Experiment Settings}
\subsection{Datasets}

We evaluate our approach on five real-world datasets used by Borisov et al.~\cite{borisov2022language}, and four datasets obtained from Open FEMA~\cite{femadatasets}. The selected FEMA datasets, namely \texttt{Public Assistance}, \texttt{Hazard Mitigation}, \texttt{Program Deliveries}, and \texttt{Program Valid Registrations}, are closely related to Trillion's business domain. Similar to Borisov et al.~\cite{borisov2022language}, the selected FEMA datasets cover a diverse range of domains, including demographic, disaster, and financial data. The number of fields in each dataset ranges from 14 to 71, totaling 222 fields.
For each dataset, we use \sysname and \rbr to synthesize 100 records for evaluation.

\subsection{Algorithm Configuration} For free-text field generation, we set the number of top-$k$ fields to $k = 10$, the number of retries to $n = 3$, and the number of samples to $s = 100$ during the preprocessing phase. We use LLaMA-70B for both field analysis and generation.

\subsection{Baselines} 
We compare \sysname with \gt, the ground truth from the Open FEMA datasets, and with \rbr. \rbr uses an LLM to read field metadata descriptions and generate data record by record. As \rbr simulates mainstream methods in which the LLM directly produces the data, its from \rbr are highly realistic, albeit at a higher cost and latency~\cite{ugare2024itergen, jung2024enhancing, guo2024generative}. To ensure a fair comparison, \rbr uses the same non-finetuned model and has access to the same metadata descriptions as \sysname. All data generation methods use the Meta LLaMA-70B model. For each method, we generate 100 records per dataset.

\section{Evaluation}\label{sec:evaluation}

\subsection{Diversity}
To quantify the diversity of the generated data, we measure vocabulary and Inter-Sample N-gram Frequency (ISNF), both of which are computational metrics used in the previous study \cite{yu2023large}.

\subsubsection{Vocabulary}
Vocabulary quantifies the overall lexical diversity by measuring the total number of unique words present in the generated content. This metric indicates whether the model produces a broad range of words or limits itself to repetitive and constrained generation.

\begin{table}[h]
\centering
\begin{tabular}{cccc}
\toprule
\textbf{Dataset} & \textbf{\sysname} & \textbf{\rbr} & \textbf{\gt} \\
\midrule
Travel Customers & 4.14 & 4.00 & 4.29 \\
Sick & 18.37 & 3.37 & 10.30 \\
HELOC & 21.83 & 7.54 & 26.42 \\
Adult & 9.40 & 8.93 & 18.67 \\
California Housing & 89.80 & 37.50 & 67.60 \\
Public Assistance & 85.64 & 29.09 & 134.64 \\
Hazard Mitigation & 50.53 & 7.26 & 50.00 \\
Program Deliveries & 67.43 & 17.74 & 60.29 \\
Program Valid Registrations & 33.30 & 7.07 & 17.56 \\
\bottomrule
\end{tabular}
\caption{Mean vocabulary scores from \sysname across fields in different datasets}
\label{tab:vocab}
\end{table}

\autoref{tab:vocab} presents the mean vocabulary scores of \sysname and the baselines across different datasets. Our method consistently generates data with a richer vocabulary than \rbr across all datasets. We also observe a significant gap in vocabulary scores between \rbr and \gt, indicating a lack of lexical diversity in the outputs produced by the \rbr method.

\subsubsection{Inter Sample N-gram Frequency}
Inter-Sample N-Gram Frequency (ISNF) quantifies the extent of redundancy across multiple samples of a field. ISNF is computed by measuring the Jaccard similarity between all pairs of generated outputs, using n-grams as the unit of comparison. This metric ranges from 0 to 1, where higher values suggest that the model produces similar outputs, and lower values reflect higher diversity.

\begin{table}[h]
\centering
\begin{tabular}{cccc}
\toprule
\textbf{Dataset} & \textbf{\sysname} & \textbf{\rbr} & \textbf{\gt} \\
\midrule
Travel Customers & 0.824 & 0.876 & 0.868 \\
Sick & 0.811 & 0.876 & 0.829 \\
HELOC & 0.512 & 0.817 & 0.691 \\
Adult & 0.638 & 0.826 & 0.784 \\
California Housing & 0.294 & 0.622 & 0.195 \\
Public Assistance & 0.316 & 0.778 & 0.289 \\
Hazard Mitigation & 0.568 & 0.825 & 0.663 \\
Program Deliveries & 0.539 & 0.809 & 0.744 \\
Program Valid Registrations & 0.738 & 0.866 & 0.856 \\
\bottomrule
\end{tabular}
\caption{ISNF scores from \sysname and baselines from different datasets}
\label{tab:isnf}
\end{table}

Mean ISNF scores reported in \autoref{tab:isnf} indicate that data distributions can vary across datasets, as shown in the \gt column. This suggest underlying distributions of real world datasets may widely fluctuated. \sysname tends to follow \gt's fluctation, especially in datasets like California Housing and Public Assistance. On the other hand, we observe \rbr have ISNF scores ranging from 0.6 to 0.9, indicating the inflexibility of the underlying distribution of generated data.

\subsection{Realism}\label{sec:eval:realism}
This section evaluates realism using various popular computational metrics \cite{borisov2022language, yu2023large}.

\subsubsection{Kullback-Leibler Divergence}
KL divergence measures the distance between discrete probability distributions. We use KL divergence to compare the distribution of data generated by \sysname to the ground truth (\gt), and contrast it with the distribution from \rbr. \autoref{tab:kl_vs_gt} reports KL divergence scores for \sysname and \rbr on numerical fields across various datasets.

\begin{table}[h]
\centering
\begin{tabular}{lrr}
\hline
\textbf{Dataset} & \textbf{$\mathrm{KL}(\text{\sysname} \Vert \text{\gt})\ \downarrow$} & \textbf{$\mathrm{KL}(\text{\rbr} \Vert \text{\gt})\ \downarrow$} \\
\hline
Public Assistance & \textbf{0.98} & 2.94 \\
Hazard Mitigation & 0.76 & \textbf{0.42} \\
Program Deliveries & \textbf{0.59} & 1.67 \\
Travel Customers & \textbf{0.36} & 0.54 \\
Sick & \textbf{0.83} & 1.54 \\
HELOC & \textbf{1.02} & 1.45 \\
Adult & \textbf{1.19} & 1.68 \\
California Housing & \textbf{1.14} & 1.49 \\
\hline
\end{tabular}
\caption{KL divergence between FASTGEN and direct-gen versus ground truth across datasets for numerical columns}
\label{tab:kl_vs_gt}
\end{table}

Overall, we observe improvements across all datasets except for Hazard Mitigation. This result suggests that \sysname is capable of generating numerical values similar to the ground truth and outperforms direct LLM-based generation. A closer inspection of the Hazard Mitigation dataset reveals that several numerical fields exhibit complex distributions that cannot be easily captured by simple mathematical formulas. In such cases, outputting only the most frequent values, as done by \rbr, results in lower error than \sysname. However, in the majority of cases, \sysname outperforms \rbr.

\subsubsection{Optimal Transport}
For categorical fields, we use the Optimal Transport (OT) distance \cite{alvarez2020geometric}. OT computes the minimum cost required to transform one probability distribution over categories into another, based on a ground cost matrix that can encode semantic similarity between categories. This makes OT particularly suitable for comparing categorical distributions where the labels may differ superficially but convey similar meanings, e.g., `True` vs `true` or `t`. KL divergence performs poorly in such cases because it treats these values as unrelated bins, which can exaggerate differences due to categorical mismatches.

\begin{table}[h]
\centering
\begin{tabular}{lrr}
\hline
\textbf{Dataset} & \textbf{$\mathrm{OT}(\text{\sysname} \Vert \text{\gt})\ \downarrow$} & \textbf{$\mathrm{OT}(\text{\rbr} \Vert \text{\gt})\ \downarrow$} \\
\hline
Public Assistance & \textbf{0.32} & 0.40 \\
Hazard Mitigation & \textbf{0.22} & 0.34 \\
Program Deliveries & \textbf{0.27} & 0.33 \\
Program Valid Registrations & \textbf{0.17} & 0.18 \\
Travel Customers & 0.07 & \textbf{0.06} \\
Sick & \textbf{0.03} & 0.09 \\
HELOC & \textbf{0.00} & 0.03 \\
Adult & \textbf{0.14} & 0.23 \\
California Housing & \textbf{0.09} & 0.32 \\
\hline
\end{tabular}
\caption{Optimal Transport distance between FASTGEN and direct-gen versus ground truth across datasets for categorical columns}
\label{tab:kl_vs_gt_categorical}
\end{table}

The report on OT distance in \autoref{tab:kl_vs_gt_categorical} shows that the data generated by \sysname is generally closer to the ground truth than the data generated by \rbr. This result suggests \sysname can effectively generate categorical and text data that mirrors the values and distribution of \gt.

\subsection{Exploratory Analysis}
We perform a deep-dive exploratory analysis on several aspects of the data to gain further insights into the performance of \sysname.

\subsubsection{Vocabulary}

\begin{figure}[h!]
    \centering
    \includegraphics[width=.8\textwidth]{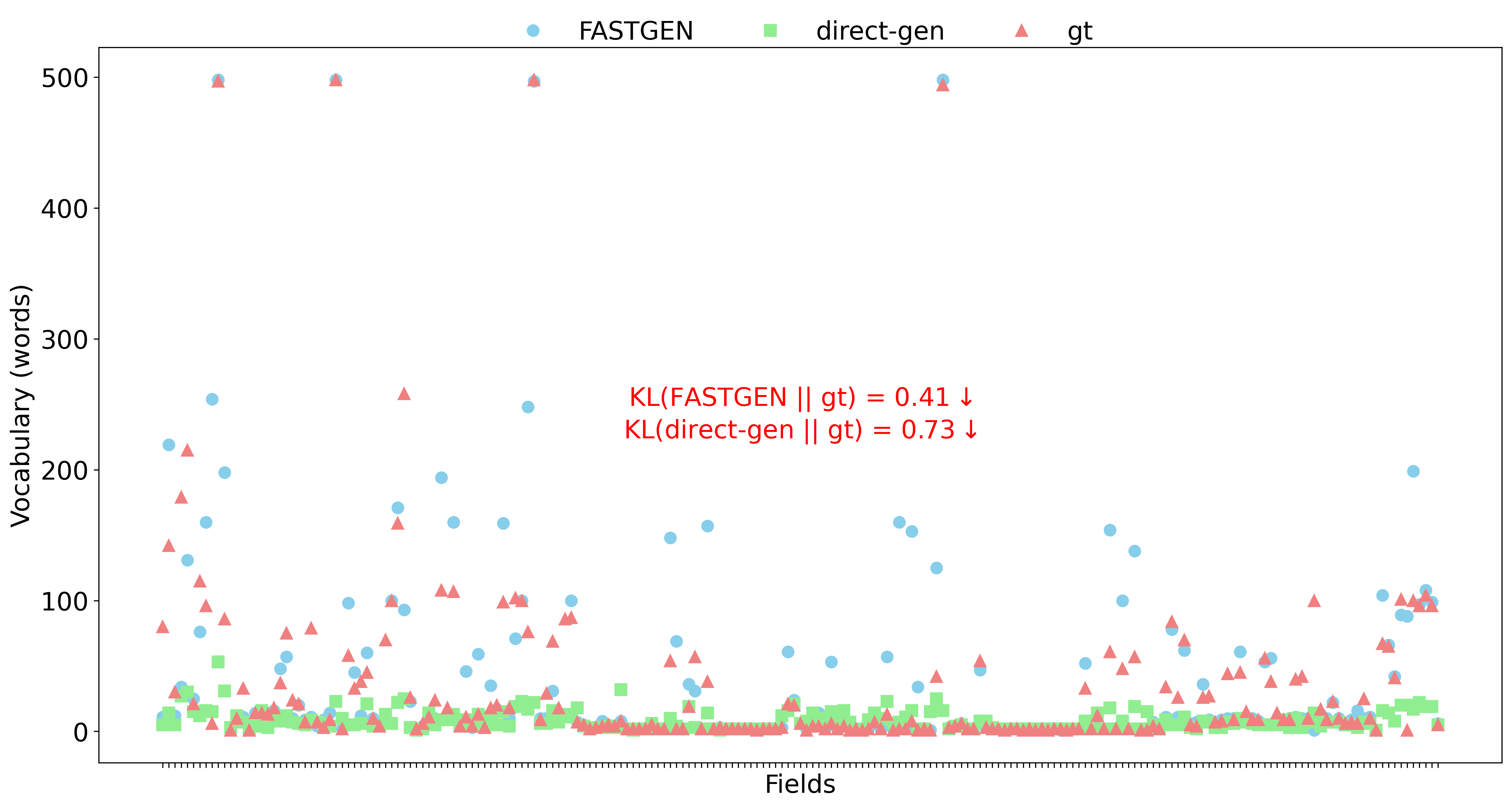}
    \caption{Vocabulary scores from \sysname and baselines across fields in four different FEMA datasets. The vocabularies of \sysname and \gt are similar, especially in columns where diversity is strictly required, such as unique IDs (columns with 500 distinct words). \rbr on the contrary, has low vocabulary in all fields.}

    \label{fig:vocab}
\end{figure}

\autoref{fig:vocab} provides a detailed view of the variation in vocabulary scores for each field in the datasets. Both \sysname and \gt achieve the highest vocabulary in four ID columns that require strictly uniform random generation. This highlights a key strength of the script generation approach used by \sysname, which cannot be achieved by \rbr. In other fields, \sysname tends to align with \gt in terms of the required vocabulary, which ranges from 0 to 300. In contrast, the \rbr method exhibits very low diversity across all fields. Further analysis shows that \rbr tends to repeat a limited set of samples with high frequency, resulting in reduced diversity.

\subsubsection{Inter Sample N-gram Frequency}

\begin{figure}[h!]
    \centering
    \includegraphics[width=.8\textwidth]{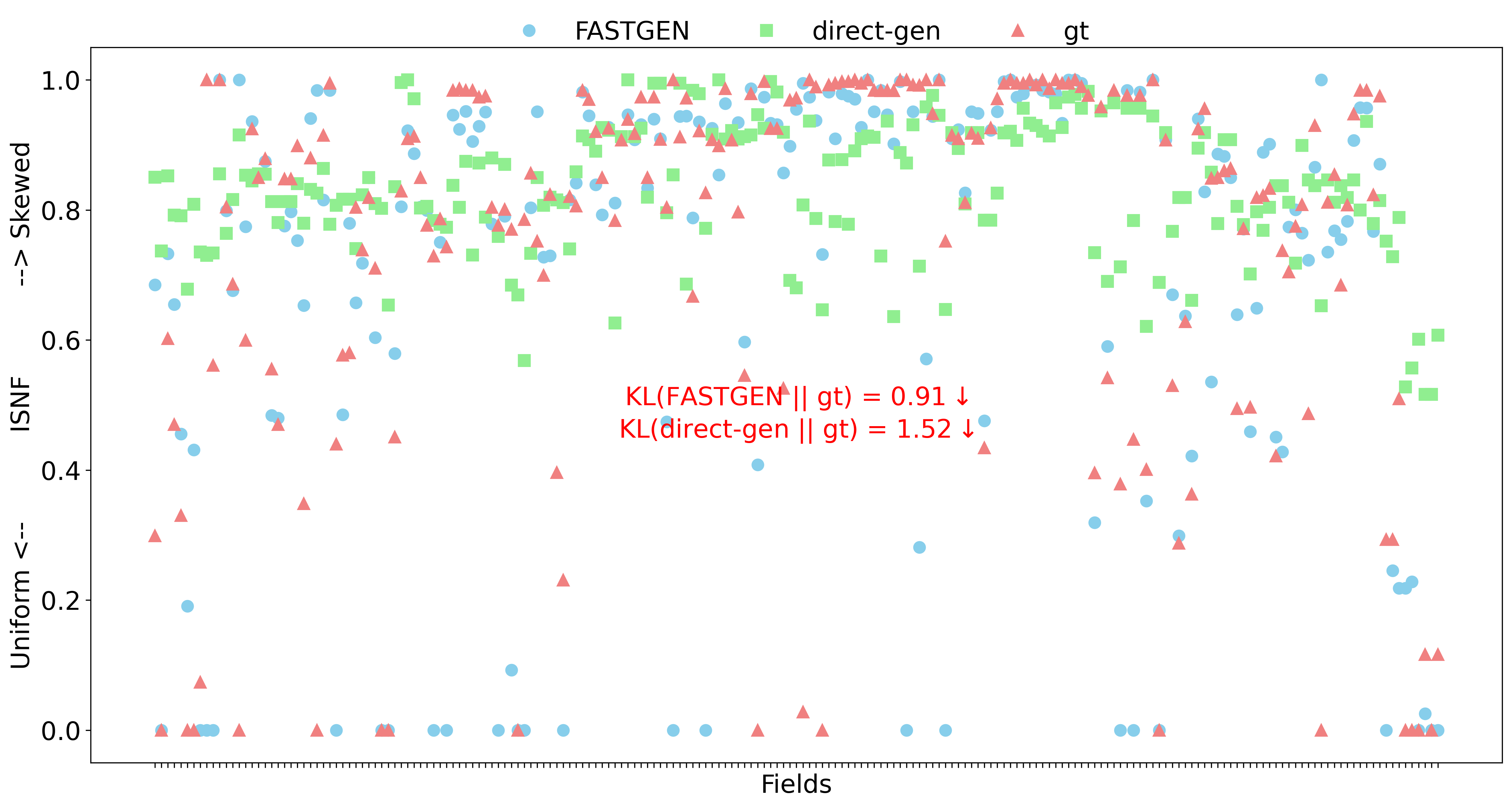}
    \caption{
    ISNF scores for \sysname and baselines across fields from different datasets. 
    The red data points show distribution shapes of ground truth data may vary widely across fields. 
    \sysname is capable of generating data with similarly fluctuating patterns}

    \label{fig:isnf}
\end{figure}

\autoref{fig:isnf} displays mean ISNF scores across fields from the datasets. ISNF quantifies the shape of a distribution, where values close to 1 indicate highly skewed distributions, and values near 0 suggest nearly uniform distributions. Most of the fields from \rbr have ISNF scores ranging from 0.6 to 0.9, indicating inflexibility and bias toward popular values. Our method and \gt, on the other hand, tend to fluctuate from field to field. This implies that \sysname can adapt to changes in the underlying distribution of each field. We also notice that our method does not completely overlap in ISNF scores with \gt, indicating that further prompt refinement is needed for alignment.

\subsection{Efficiency Analysis}
To evaluate the efficiency of \sysname, we measure the number of tokens generated by the LLM to produce a set of records. The number of generated tokens is a direct indicator of efficiency, as it is linearly correlated with both the time and inference cost of an LLM.

\begin{figure}[h!]
    \centering
    \includegraphics[width=.6\textwidth]{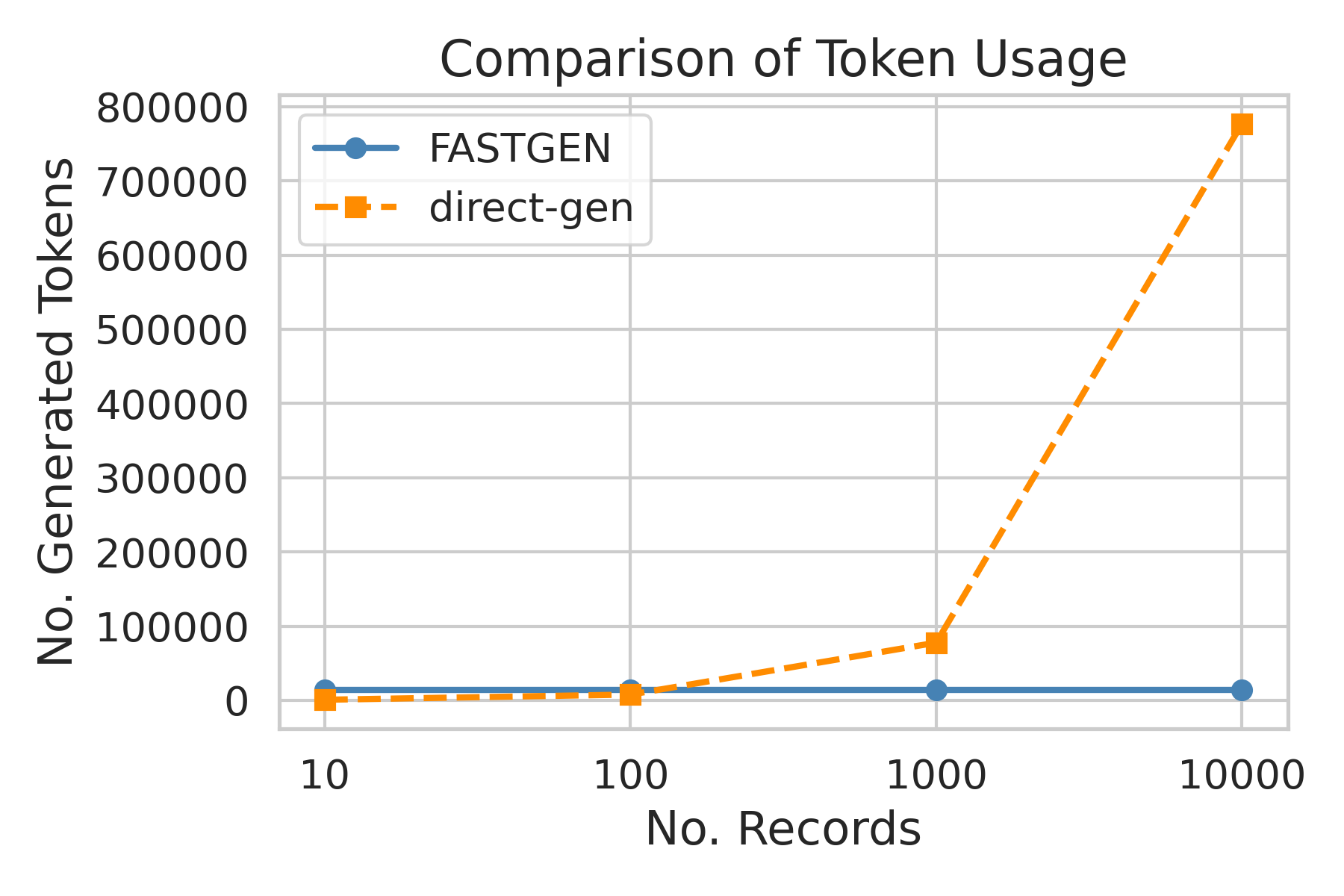}
    \caption{Token generation performance for \sysname for producing the target number of records.}
    \label{fig:throughput}
\end{figure}

\autoref{fig:throughput} shows the number of tokens required by the LLM to generate outputs for both \sysname and \rbr across different target record counts. Since \rbr generates tokens directly for each record, its token usage increases rapidly, reaching up to 800,000 tokens to produce 10,000 records for the Travel Customers dataset. In comparison, \sysname requires approximately 6× fewer tokens for 1,000 records and 60× fewer tokens for 10,000 records.

\autoref{tab:costanalysis} illustrates the practical implications in terms of cost and latency. We consider a Llama-70B model hosted on Azure, which charges \$0.71 per million output tokens and generates tokens at a speed of 55 tokens per second. Under this setting, generating 100,000 records with \rbr on the Travel Customers dataset would take 40 hours and cost \$55. In contrast, \sysname requires only 0.07 hours and costs just \$0.096.

\begin{table}[h]
\centering
\begin{tabular}{lccc}
\hline
\textbf{Method / Metric} & \textbf{1{,}000 Recs} & \textbf{10{,}000 Recs} & \textbf{100{,}000 Recs} \\
\hline
\sysname Cost (\$)  & 0.096 & 0.096 & 0.096 \\
\sysname Time (h)   & 0.07  & 0.07  & 0.07 \\
\rbr Cost (\$)      & 0.55  & 5.50  & 55.00 \\
\rbr Time (h)       & 0.40  & 4.00  & 40.00 \\
\hline
\end{tabular}
\caption{Estimated cost and end-to-end latency (hours) for LLaMA-70B on Azure to generate data for \texttt{Travel Customers} dataset}
\label{tab:costanalysis}
\end{table}

\section{Lessons, Limitation and Future Works}
\subsection{Human Feedback for Continuous Improvement}
Our evaluation on realism (\autoref{sec:eval:realism}) reveals a gap between the generated and ground truth data, primarily due to the limited expressiveness of field metadata descriptions. As a result, the generated data may require iterative refinement to achieve the desired quality. To address this challenge, we aim to enhance transparency in our generation process by exposing the LLM’s reasoning during the categorization step and providing access to the generated sampling script. This allows end users to directly modify the script to better align with their requirements or supply additional information about the data distribution, ensuring greater control over the final output.

 
\subsection{Insights on LLM Behavior}
Our analysis provides several insights into how LLMs behave when tasked with generating and evaluating synthetic data. Although LLMs can be powerful automated tools, they can make mistakes, e.g., produce buggy scripts, or failure to fully interpret complex field metadata requirements. We found that using larger models with more robust reasoning capabilities can partially mitigate these issues.

Another key lesson is when requirements are under-specified or highly complex, it can be advantageous to keep prompts simple, thereby granting the LLM latitude in devising creative solutions. In the case of free-text field generation, the LLM may choose different strategies based on data complexity. In scenarios such as generating city names, the LLM might rely on established libraries such as \emph{Faker} \cite{faker} for a realistic and diverse solution. Conversely, when relevant tools are not readily available such as for street addresses, the LLM falls back on heuristic rules that combine street numbers with street names.

\subsection{Limitation on Cross-field Relationship}
A drawback of our current solution is the independent generation of fields, which overlooks relationships across columns. For instance, selecting a city name without considering the associated state or country can diminish the realism of the final dataset. In future work, we plan to address this limitation by using LLMs to capture the dependencies among fields. Values of the secondary fields, which are dependent on primary fields, could be generated programmatically based on the captured dependencies. The main challenge is that building a complete dependency map for all fields becomes exponentially costly as the number of fields increases. Instead, we aim to identify the most important dependencies by first asking an LLM to highlight the key fields on which others depend. We then expand from these primary fields to identify a limited set of secondary fields. This approach captures the most significant dependencies, thereby improving the realism of our method.

\section{Conclusion}
We proposed a cost-effective approach for generating large-scale synthetic tabular data by leveraging LLMs to infer underlying data distributions rather than generating records individually. By categorizing fields into numerical, categorical, and free-text types, our method ensures adaptability across diverse domains. Experimental results demonstrate that our distribution-based strategy produces more diverse and realistic data while substantially reducing computational overhead and inference costs compared to directly using LLMs for generation.

At Trillion, we plan to apply this methodology to streamline test dataset generation, mitigating data bottlenecks in production systems and accelerating development cycles. We share key insights from developing this methodology to support researchers and practitioners seeking scalable, high-quality synthetic data generation. Future work can focus on refining distribution inference techniques and extending this method to more complex data structures.

\bibliographystyle{plain}  
\bibliography{ref}

\end{document}